\documentclass[runningheads]{llncs}

\usepackage{eccv}

\usepackage{eccvabbrv}

\usepackage{graphicx}
\usepackage{booktabs}
\usepackage{amsmath}
\usepackage{amssymb}
\usepackage{array}
\usepackage{caption}
\usepackage{multirow}
\usepackage{makecell}
\usepackage{siunitx}
\usepackage{pifont}
\usepackage{colortbl}
\usepackage{enumitem}
\usepackage[normalem]{ulem}
\usepackage{adjustbox}
\usepackage{cellspace}
\usepackage{float}
\usepackage{amssymb}
\usepackage{booktabs}
\usepackage{wrapfig}
\usepackage{hhline}
\usepackage[dvipsnames]{xcolor}
\usepackage{footmisc}

\usepackage[accsupp]{axessibility}  % Improves PDF readability for those with disabilities.

\usepackage{algorithm}
\usepackage{algorithmic}
\usepackage{eqparbox}

\newcommand{\cmark}{\ding{51}}%
\newcommand{\xmark}{\ding{55}}%

\usepackage{hyperref}

\usepackage{orcidlink}

\begin{document}

% ---------------------------------------------------------------
% TEASER
% \input{Figures/teaser}

% TODO REVIEW: Replace with your title
\title{Receler: Reliable Concept Erasing of Text-to-Image Diffusion Models via\\Lightweight Erasers}

% TODO REVIEW: If the paper title is too long for the running head, you can set
% an abbreviated paper title here. If not, comment out.
\titlerunning{Receler}

% TODO FINAL: Replace with your author list. 
% Include the authors' OCRID for the camera-ready version, if at all possible.

\author{
    {Chi-Pin Huang}%
    \thanks{Equal Contribution}%
    \inst{1,\dagger}\orcidlink{0009-0003-7738-3054} \and
    {Kai-Po Chang}%
    \textsuperscript{\textasteriskcentered}%
    \inst{1}\orcidlink{0009-0008-5182-2019} \and
    Chung-Ting Tsai\inst{2}\orcidlink{0009-0009-0877-5795} \and \\
    Yung-Hsuan Lai\inst{2}\orcidlink{0009-0006-6194-4189} \and
    Fu-En Yang\inst{3}\orcidlink{0000-0003-0102-7101} \and
    Yu-Chiang Frank Wang\inst{1,3,\ddagger}\orcidlink{0000-0002-2333-157X}
}

% TODO FINAL: Replace with an abbreviated list of authors.
\authorrunning{Huang et al.}
% First names are abbreviated in the running head.
% If there are more than two authors, 'et al.' is used.

% TODO FINAL: Replace with your institution list.
\institute{
    Graduate Institute of Communication Engineering, National Taiwan University
    \and
    National Taiwan University, $^3$ NVIDIA\\
    $^\dagger$ \email{f11942097@ntu.edu.tw}, $^\ddagger$ \email{frankwang@nvidia.com}
}

\maketitle

\begin{abstract}

Concept erasure in text-to-image diffusion models aims to disable pre-trained diffusion models from generating images related to a target concept. To perform reliable concept erasure, the properties of robustness and locality are desirable. The former refrains the model from producing images associated with the target concept for any paraphrased or learned prompts, while the latter preserves its ability in generating images with non-target concepts. In this paper, we propose \textbf{Re}liable \textbf{C}oncept \textbf{E}rasing via \textbf{L}ightweight \textbf{Er}asers (\textbf{Receler}). It learns a lightweight Eraser to perform concept erasing while satisfying the above desirable properties through the proposed concept-localized regularization and adversarial prompt learning scheme. Experiments with various concepts verify the superiority of Receler over previous methods.  Code is available at \url{https://github.com/jasper0314-huang/Receler}.

\end{abstract}

\begin{figure}[!t]
    \setlength{\linewidth}{\textwidth}
    \setlength{\hsize}{\textwidth}
    \centering
    \includegraphics[width=0.975\linewidth]{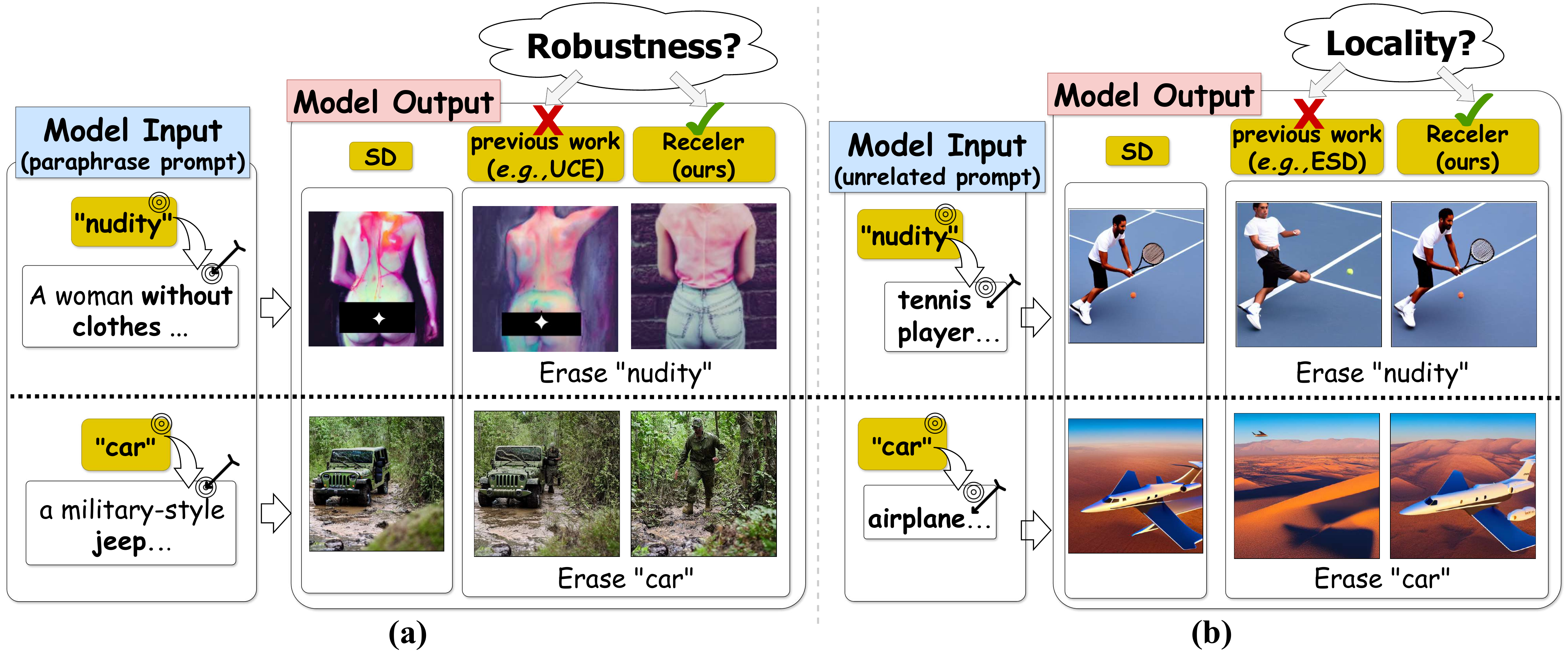}

    \caption{\textbf{Illustration of (a) robustness and (b) locality-preserving in concept erasing.} The former requires models to be robust against paraphrased attacks from the target concept (\ie, ``nudity''), while the latter aims to preserve the visual content of non-target concepts (\eg, ``tennis player'' or ``airplane''). Note that SD denotes Stable Diffusion~\cite{rombach2021compvissd}; UCE~\cite{gandikota2023unified} and ESD~\cite{gandikota2023erasing} are recent works on concept erasing.}

    \label{fig:teaser}
\end{figure}
\section{Introduction}
\label{sec:intro}

In recent years, text-to-image generation models~\cite{ramesh2022dalle2, saharia2022imagen, balaji2022ediffi, rombach2021compvissd, nichol2021glide} have achieved unprecedented success in generating photo-realistic images which benefit various industrial applications~\cite{ramesh2022dalle2, saharia2022imagen}. Despite their apparent success and convenience, these models may produce images that are deemed NSFW (Not Safe for Work)~\cite{hunter2023ai} or infringe upon intellectual property and portrait rights~\cite{andersen2023stabilityailtd}, \eg, generating nudity, violent content, or imitating the style of well-known artists. This issue is mainly due to the memorization of the large-scale training data sourced from the web~\cite{somepalli2023diffusion, carlini2023extracting}. To address the above problem, an intuitive solution is to manually filter out the inappropriate images and re-train the model. However, as pointed out in~\cite{gandikota2023unified}, this may lead to unpredictable results such as exposing more inappropriate content to be memorized~\cite{carlini2022privacy} or incomplete visual concept removal~\cite{oconnor2022stable2.0}. Moreover, even though filtered image data can be collected, re-training generative models is still computationally expensive~\cite{kumari2023ablating, gandikota2023erasing, sinitsin2020editable}.

Instead of processing training data and re-training models, an alternative is to \textit{erase} or \textit{unlearn} specific concepts from the pre-trained model~\cite{gandikota2023erasing, kumari2023ablating, zhang2023forget, gandikota2023unified}. That is, given a concept described in text, the pre-trained model is fine-tuned to forget that concept so that the associated image content cannot be generated from the fine-tuned model. In practice, it would be desirable to perform concept erasing with sufficient \textit{reliability}, which suggests two desirable properties: \textit{locality} and \textit{robustness}. Locality indicates the ability to preserve the model generalization in synthesizing content not associated with the target concept~\cite{ni2023ores, sinitsin2020editable}. Robustness requires erased models to effectively remove the target concept~\cite{ni2023ores, sinitsin2020editable}, while not be circumvented by paraphrased prompts that aim to recover the target concept (e.g., ``car'' vs. ``jeep''). While both locality and robustness have been recently discussed in the field of NLP~\cite{de2021editing, yao2023editing, huang2023transformer, meng2022locating, chenyang2023eul}, the developed techniques cannot be directly applied to text-to-image generative models for performing concept erasing.

Recently, a number of methods for concept erasing or unlearning for diffusion models have been proposed~\cite{gandikota2023erasing, kumari2023ablating, zhang2023forget, gandikota2023unified}. For example, Ablating~\cite{kumari2023ablating} predefines an anchor concept for each target concept that needs to be unlearned and then achieves model unlearning by mapping the image distribution of the target concept to that of the anchor concept, \eg mapping ``grumpy cat'' to ``cat.''
Inspired by classifier-free guidance~\cite{ho2022cfg}, ESD~\cite{gandikota2023erasing} fine-tunes the model to predict negatively guided noise. In other words, it decreases the probability of generating images of the target concept, thus unlearning that concept. FMN~\cite{zhang2023forget} designs a computationally efficient unlearning method by directly minimizing the cross-attention weights corresponding to the target concept in the input text prompt, encouraging the model to ignore the concept. UCE~\cite{gandikota2023unified} employs a closed-form editing approach to optimize the projection matrices of keys and values in cross-attention layers, ensuring the model maintains locality when unlearning the target concept.

Although promising progress has been made in erasing specific target concepts, most existing works are not specifically designed to preserve model \textit{locality} and \textit{robustness}. For example, Ablating~\cite{kumari2023ablating} and ESD~\cite{gandikota2023erasing} fine-tune a considerable amount of parameters within pre-trained diffusion models to achieve concept erasure, which inevitably compromises the original capabilities of the model.
On the other hand, methods such as UCE~\cite{gandikota2023unified} and FMN~\cite{zhang2023forget} only modify specific parameters (\ie, the projection matrices of keys and values in cross-attention layers) responsible for encoding input textual features instead of visual ones.
These methods would be vulnerable to rephrased target concepts since they only learn to dissociate textual prompts in the cross-attention layers (\ie, lack of ability to recognize that paraphrased queries are semantically similar to the erased target concept). For example, a diffusion model that has been erased of the concept of ``car'' may still produce images of jeeps due to its inability to recognize that jeeps fall under the category of cars; hence, it cannot remove the attention to jeep in cross-attention layers.
As a result, proposing a concept-erasing method that addresses \textit{locality} and \textit{robustness} continues to pose a crucial challenge.

In this paper, we propose \textbf{Re}liable \textbf{C}oncept \textbf{E}rasing via \textbf{L}ightweight \textbf{Er}asers (\textbf{\textit{Receler}}) for erasing concepts from pre-trained diffusion models, exhibiting sufficient \emph{locality} and \emph{robustness} properties. \textit{Receler} involves a lightweight eraser (only 0.37\% of the U-Net parameters), which is designed to remove a target concept from the outputs of cross-attention layers. During this unlearning process, we train the eraser while preserving the image generation capability of the pre-trained diffusion models. Furthermore, a concept-localized regularization is introduced to ensure that the eraser focuses on erasing the target concept. This regularization prevents the generation of non-target concepts from being affected, thereby preserving \textit{locality}. As for \textit{robustness}, we advance adversarial prompt learning, which optimizes the adversarial prompts that induce the model to generate images of the target concept and then fine-tunes the eraser to erase images generated with these prompts. By training our eraser with concept-localized regularization and adversarial prompt learning, we are able to preserve the image generation capability of non-target concepts and robustly refrain the model from generating images with target-concept content.

We now summarize the contributions of this work below:
\begin{itemize}[leftmargin=1.6em]
    \setlength\itemsep{0.3em}
    \item We present \textbf{Re}liable \textbf{C}oncept \textbf{E}rasing via \textbf{L}ightweight \textbf{Er}asers (\textbf{\textit{Receler}}), a novel approach using a lightweight eraser (only 0.37\% of the U-Net parameters) for reliable and efficient concept erasing.
    
    \item Locality is introduced through concept-localized regularization, which constrains the eraser for precise erasing of the target concept without affecting the generation of non-target ones.
    
    \item \textit{Receler} is trained against adversarial prompts, imitating paraphrased prompts of target concepts, resulting in improved robustness in concept erasure.

\end{itemize}
\section{Related Works}
\label{sec:related_works}

\subsection{Erasing Concepts from Diffusion Models}
Text-to-image diffusion models~\cite{balaji2022ediffi, rombach2021compvissd, ramesh2022dalle2, saharia2022imagen, nichol2021glide} have been shown to generate high-quality images with impressive generalization.
However, such models are typically trained on extensive web-crawled data (e.g., LAION-5B~\cite{schuhmann2022laion5b}), which could memorize NSFW or copyrighted content, leading to the generation or replication of undesired images~\cite{carlini2023extractingdm, somepalli2023datareplication}.
To address this issue, some works explore the solution without the need to update the model weights. For instance, Stable Diffusion~\cite{rombach2021compvissd} employs an unsafe content classifier to filter out risky outputs, while SLD~\cite{schramowski2023sld} uses negative guidance to prevent inappropriate content generation. However, the former relies on the reliability of pre-trained classifiers, while the latter only suppresses undesired concepts without complete removal.

In response, several works focus on fine-tuning the diffusion model to erase the target concepts~\cite{gandikota2023erasing, kumari2023ablating, zhang2023forget, gandikota2023unified}. Ablating~\cite{kumari2023ablating} associates each concept to be erased, \eg, ``grumpy cat,'' with a broader, predefined anchor concept, \eg, ``cat'' and fine-tunes the diffusion model to map the generated image of the target concept to that of the anchor concept by minimizing the L2 distance of predicted noises. Inspired by classifier-free guidance~\cite{ho2022cfg}, ESD~\cite{gandikota2023erasing} proposes to decrease the likelihood of generating images belonging to the target concept. This is achieved by fine-tuning the diffusion model to predict negatively guided noises, effectively steering the model's conditional prediction away from the erased concept. FMN~\cite{zhang2023forget} adopts attention resteering, a computationally efficient unlearning method, to identify attention maps associated with the target concept in the diffusion U-Net's cross-attention layers. By minimizing the attention weights corresponding to the target concept, the diffusion model gradually disregards the target concept during image generation, facilitating the erasure of the concept. UCE~\cite{gandikota2023unified} employs a closed-form editing method to optimize the projection matrices of keys and values in cross-attention. The objective is to align the embedding of a source prompt (\eg, ``a photo of an airplane'') more closely with that of a destination (\eg, an empty string), while leaving other unrelated concepts unchanged.
Despite their effectiveness in erasing concepts, most current methods are not particularly designed to preserve robustness against paraphrased prompts (\eg, ``nudity'' vs. ``without clothes''). Moreover, both~\cite{kumari2023ablating} and~\cite{gandikota2023erasing} require fine-tuning a considerable number of model parameters, which might affect the model capability and consequently compromise the locality property.

\subsection{Controlling Text-to-Image Diffusion Models}
Parameter-Efficient Fine-Tuning (PEFT) is a training scheme that addresses the challenges of extensive parameter updates, especially for large language models. These approaches update only a small subset of parameters, thereby reducing the risk of compromising the pre-trained capabilities of the model. Recent researches~\cite{gal2022textualinversion, ruiz2023dreambooth, kumari2023customdiffusion, mou2023t2iadapter, zhang2023controlnet, ye2023ipadapter, zhao2023unicontrolnet} have applied PEFT to control text-to-image diffusion models.
For instance, some studies~\cite{gal2022textualinversion, ruiz2023dreambooth, kumari2023customdiffusion} empower the model to learn personalized or unseen concepts by fine-tuning a new text token and a small number of parameters using few user-provided exemplar images.
Meanwhile, other works~\cite{mou2023t2iadapter, zhang2023controlnet, ye2023ipadapter, zhao2023unicontrolnet} aim to enable diffusion models to generate images based on additional conditions, \eg, edge maps, depth maps, or segmentation masks. They fine-tune lightweight task-specific modules with condition-image pairs to achieve control over the image generation process.
Despite the effectiveness of these methods, they achieve learning new concepts by accessing the associated data of interest during training. As for concept erasing, since one only observes the description of the concept to be unlearned, existing PEFT-based methods cannot be directly applied.

\subsection{Adversarial Attack \& Training}
In adversarial attacks~\cite{dong2018mifgsm, kurakin2016ifgsm, goodfellow2014fgsm, akhtar2021adversarialsurvey}, adversarial examples are deliberately constructed inputs, which would deceive models into making incorrect predictions. Popular methods such as Fast Gradient Sign Method (FGSM~\cite{dong2018mifgsm}) and its variants (I-FGSM~\cite{dong2018mifgsm} and MI-FGSM~\cite{dong2018mifgsm}) targeted at attacking classification models, utilizing the resulting loss gradients to produce imperceptible perturbations and induce misclassification. In contrast, recent approaches~\cite{chin2023p4d, tsai2023ring, zhang2023togenornot} introduce prompt-based adversarial attacks tailored to provoke seemingly unlearned diffusion models into generating images of the unlearned concepts. For instance, P4D~\cite{chin2023p4d} learns prompts to reconstruct noise associated with the target concept in diffusion models, and Ring-A-Bell~\cite{tsai2023ring} extracts holistic concept representations from CLIP model~\cite{radford2021clip} to generate model-agnostic attack prompts. These learned attack prompts have been shown to provoke unlearned models to regenerate images of the erased concept, posing potential issues in text-to-image generative models.

To defend against such adversarial attacks, various adversarial training strategies have been proposed~\cite{goodfellow2014fgsm, madry2017advrstrain, bai2021recentadvrs}. Such adversarial training schemes expose models to adversarial examples during training, enabling the models to learn and recognize these examples. Consequently, these adversarially trained models can respond accurately when encountering adversarial examples during inference. Inspired by adversarial training approaches, we employ an adversarial erasing learning scheme to introduce additional robustness to the unlearned model. We will detail our proposed framework in the following section.

\section{Method}

\subsubsection{Problem formulation.}
We first define the setting and notations of our \textit{Receler}. Given a pre-trained text-to-image diffusion model, parameterized by $\theta$, we aim to erase a textual concept $c$ from the model without requiring access to the corresponding image data. The erasure is considered successful when the model no longer generates images that contain or represent the concept $c$ (\eg, ``nudity''-erased model should not generate any images with exposed body parts). 

\begin{figure*}[t!]
  \centering
  \includegraphics[width=0.975\textwidth]{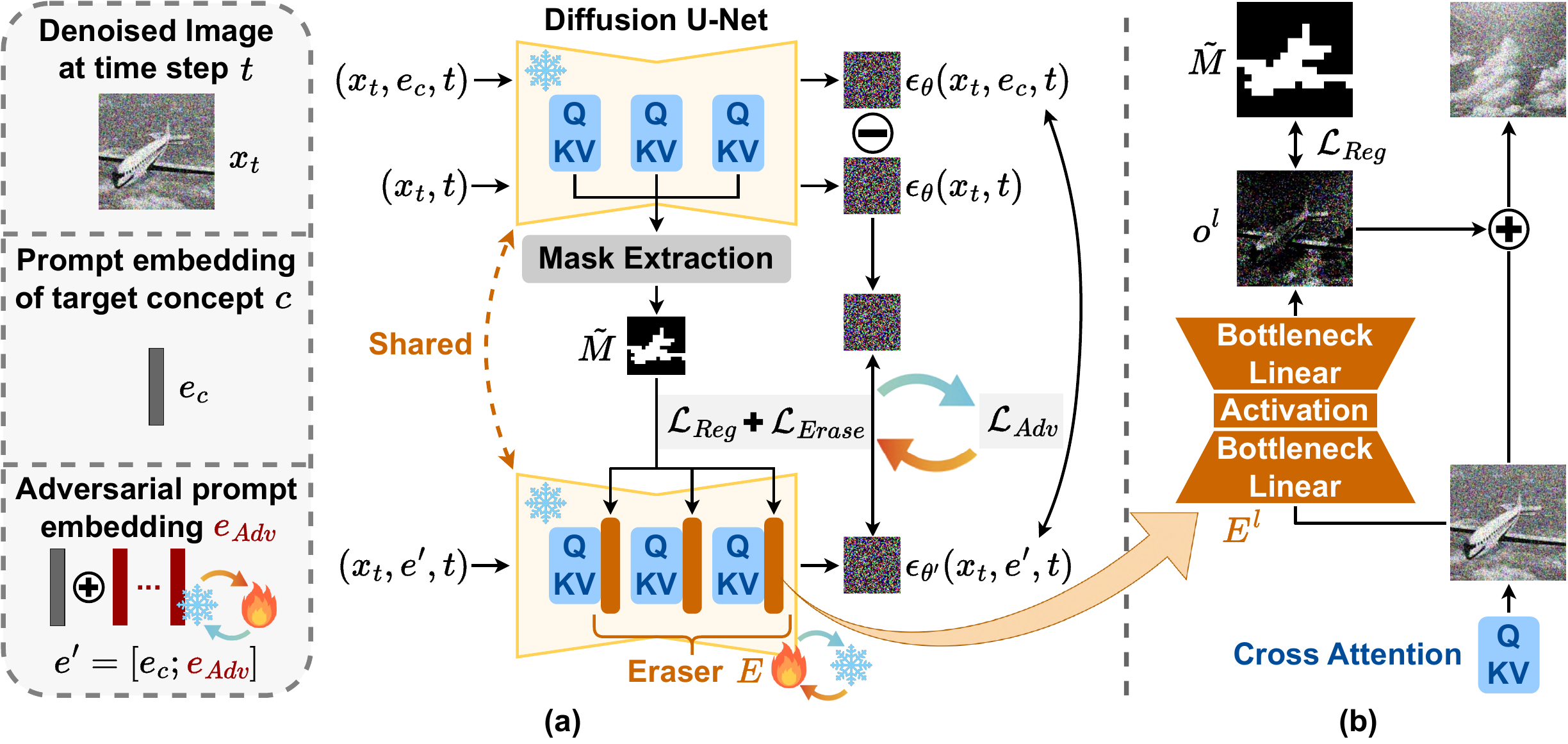}

  \caption{
  \textbf{Overview of \textit{Receler}.} (a) \textit{Receler} involves iterative learning of a lightweight Eraser $E$ and adversarial prompt embedding $e_{\textit{Adv}}$. The former is trained to erase the target concept $c$ while preserving non-target concepts, and the latter learns to imitate the prompts to recover visual content associated with the concept previously erased. (b) The Eraser $E$ is inserted after each cross attention layer of Diffusion U-Net to remove the target concept from its outputs, with prediction $o^l$ directly added to the cross attention output.
  }
  \label{fig:main}

\end{figure*}

As depicted in~\cref{fig:main}, our method employs a lightweight Eraser $E$, parameterized by $\theta_{E}$, to learn to erase the target concept $c$, as discussed in~\cref{ssec:eraser}. To introduce the desirable locality and robustness to our model, we incorporate concept-localized regularization and adversarial prompt learning schemes into our framework, as detailed in~\cref{ssec:regularization} and~\cref{ssec:adversarial}.

\subsection{Concept Erasing with Lightweight Eraser}\label{ssec:eraser}
In order to erase particular visual concepts from a pre-trained diffusion model, we introduce a lightweight adapter-based eraser. As depicted in~\cref{fig:main}, by solely fine-tuning the newly introduced $\theta_E$, our goal is to unlearn the target visual concept while preserving model generalization on non-target concepts. To be specific, our eraser is designed to \emph{remove} the target concept from the output visual features of each cross-attention layer within the diffusion U-Net. This is based on the fact that these layers are responsible for incorporating the input text concept into the visual features. Thus, our eraser is positioned subsequent to each of the cross-attention layers, which is in line with the empirical analysis presented in~\cite{xiang2023closer}.
To erase the concept, the eraser is trained to predict the negatively guided noises~\cite{ho2022cfg, gandikota2023erasing} that move the model's prediction away from the erased concept. The objective is defined as:
\begin{equation}\label{eq:esd_loss}
\begin{split}
\mathcal{L}_{\textit{Erase}} &= \mathbb{E}_{x_t,t} \left[ \lVert \epsilon_{\theta'}(x_t, e_c, t) - \epsilon_{E} \rVert^{2} \right], \\
\text{where}\ \epsilon_E &=\ \epsilon_{\theta}(x_t, t) - \eta \left[ \epsilon_{\theta}(x_t, e_c, t) - \epsilon_{\theta}(x_t, t) \right].
\end{split}
\end{equation}

\noindent Note that $\theta' = \{\theta, \theta_{E}\}$ represents the parameters of the diffusion model plugged with eraser, $x_t \in \mathbb{R}^{W \times H \times d}$ is the denoised image at timestep $t$ sampled from $\theta'$ conditioned on $c$, $e_c$ is the text embedding of concept $c$, and $\epsilon_E$ is the negatively guided noises~\cite{gandikota2023erasing, ho2022cfg} predicted by $\theta$.
By minimizing the L2 distance between $\epsilon_{\theta'}(x_t, e_c, t)$ and $\epsilon_E$, the eraser learns to reduce the probability of the generated image $x$ belongs to the target concept $c$, thus effectively erasing the concept.

\subsection{Concept-Localized Regularization for Erasing Locality}\label{ssec:regularization}

As the first desirable property in concept erasing, locality refers to preserving the model's ability to synthesize content unrelated to the target, which is realized by enforcing the eraser to affect the image synthesis process only if the target concept is present. To achieve this, we introduce concept-localized regularization into our \textit{Receler} by leveraging the spatial information associated with
the target concept’s text tokens to regularize the eraser outputs.
Specifically, inspired by~\cite{tang2022daam, cao2023masactrl}, we obtain the binary target concept mask $M \in \mathbb{R}^{W/4 \times H/4}$ by thresholding the attention maps when predicting $\epsilon_\theta(x_t, e_c, t)$ as follows:
\begin{equation}\label{eq:mask_extraction}
M_{i,j} = 
\left\{
\begin{aligned}
&1, && \text{if }\ \frac{1}{\left| \mathcal{S} \right|} \sum_{s \in \mathcal{S}} A^{s}_{i,j} \geq \tau, \\
&0, && \text{otherwise}.
\end{aligned}
\right.
\end{equation}
where $\mathcal{S}$ is the set of indices of all U-Net's mid-layers with resolution $(\frac{W}{4}, \frac{H}{4})$, $A^{s} \in \mathbb{R}^{W/4 \times H/4}$ is the cross-attention map in the $s$-th layer of the text tokens corresponding to the concept $c$, and $\tau$ is a pre-defined threshold. With $M$ obtained, we calculate the following regularization loss to regularize the outputs of the eraser as follows:
\begin{equation}\label{eq:loss_reg}
\mathcal{L}_{\textit{Reg}} = \frac{1}{L} \sum_{l=1}^{L} \lVert o^{l} \odot (1-\Tilde{M}) \rVert^{2},
\end{equation}
where $L$ is the number of U-Net's layers, $\odot$ is the element-wise product, $o^l \in \mathbb{R}^{w^l \times h^l \times d}$ is the output of the eraser in the $l$-th layer with the resolution ($w^l$, $h^l$), and $d$ is the feature dimension. Note that $\Tilde{M}$ is the shorthand of $M$ being bicubically upscaled to the same resolution as $o^{l}$. With the above regularization introduced, our \textit{Receler} is enforced to preserve the generation of non-target concepts. Thus, the diversity and fidelity of the original model can be maintained.

\subsection{Adversarial Prompt Learning for Erasing Robustness}\label{ssec:adversarial}

To further ensure our \textit{Receler} being robust against prompting attacks (\eg, by paraphrased or learned prompts), we introduce an adversarial learning strategy into our framework to enrich such model robustness.
In our framework, we present a unique \textit{adversarial prompt learning} scheme, which learns prompting attacks that would induce the diffusion model to synthesize the images containing previously erased concepts. To achieve this, we design the adversarial loss, which optimizes the continuous soft prompts $e_{\textit{Adv}}$ by encouraging such learned prompts to imitate the malicious prompts, as illustrated in~\cref{fig:main}. More precisely, the objective $\mathcal{L_\textit{Adv}}$ is defined as follows:
\begin{equation}\label{eq:loss_adversarial}
    \begin{split}
        \mathcal{L}_\textit{Adv} = \mathbb{E}_{x_t,t} \left[ \lVert \epsilon_{\theta'}(x_t, e', t) - \epsilon_{M} \rVert^{2} \right],
    \end{split}
\end{equation}
where $e' = [e_c;e_\textit{Adv}]$ is the concatenated prompts of the erased target concept embedding and the learned soft prompts, and $\epsilon_{M} = \epsilon_{\theta}(x_t, e_c, t)$ represents the malicious noise predicted by the pre-trained diffusion model (without safety mechanism) conditioned on the target concept. By minimizing this adversarial loss, $e_{\text{Adv}}$ learns to regenerate the unlearned concept from the erased model. The optimization of the soft prompt $e_{\textit{Adv}}$ and the eraser $\theta_E$ is performed iteratively, with each being fixed while the other is trained. Thus, $\theta_E$ and $e_{\textit{Adv}}$ are trained against each other to improve model robustness. For more details, please refer to the pseudo-algorithm in the supplementary material.
\section{Experiments}
In this section, we first conduct quantitative experiments to assess the robustness and locality of \textit{Receler} compared to state-of-the-art baselines, followed by ablation studies of our method. Lastly, we present qualitative comparisons and visualizations to demonstrate its effectiveness.

\subsubsection{Datasets.}\label{ssec:datasets}
We conduct experiments on erasing objects defined in the CIFAR-10 dataset~\cite{krizhevsky2009cifar10} and on erasing inappropriate contents from the Inappropriate Image Prompts (I2P) dataset~\cite{schramowski2023sld}:
\begin{itemize}[leftmargin=1.6em]
    \setlength\itemsep{0.3em}
    \item \textbf{Object Erasure}.
    To evaluate the effectiveness of erasure methods in erasing common visual concepts, we choose to erase ten class labels from CIFAR-10~\cite{krizhevsky2009cifar10}. Note that during our experiment, we only utilize the label set, not the images. For comprehensive assessment, we devise two types of evaluation prompts for each class: Firstly, we use simple prompts formatted as ``A photo of \{class\}'' to evaluate the efficacy of erasure in removing the target concept. Secondly, to further assess the robustness, we generate 50 paraphrased prompts for each class using ChatGPT\footnote{https://chat.openai.com/} to simulate real-world scenarios where prompts are typically more complex and target concepts may not be explicitly mentioned. For example, a paraphrased prompt for ``airplane'' is ``A sleek, black stealth bomber flying low over a desert landscape at dusk.'' More details can be found in the supplementary material.
    
    \item \textbf{Inappropriate Content Erasure}. The I2P dataset~\cite{schramowski2023sld}, collected from a text-to-image prompt dataset DiffusionDB~\cite{wang2022diffusiondb}, comprises 4,703 real-world, user-generated prompts that produce inappropriate images, including hate, harassment, violence, self-harm, shocking, sexual, and illegal content.

\end{itemize}

\subsubsection{Evaluation Setup.}\label{ssec:setup}
We assess the robustness and locality of the erasure methods for object erasure and inappropriate content erasure as follows:
\begin{itemize}[leftmargin=1.6em]
\item \textbf{Object Erasure}. 
For each method, we fine-tune ten models, each erasing one CIFAR-10 class. Each model is then evaluated by: 1) Efficacy ($\text{Acc}_E$): the percentage of the target class being erased when inputting simple prompts; lower values are better. 2) Robustness ($\text{Acc}_R$): the percentage of the target class being erased when inputting paraphrased prompts; lower values are better. 3) Locality ($\text{Acc}_L$): the percentage of non-target classes being preserved; higher values are better.
To assess efficacy and robustness, we generate 150 images using simple and paraphrased prompts for the \emph{target class}, respectively. For Locality, we generate 50 images for each of the nine \emph{non-target} classes using paraphrased prompts.
We then use GroundingDINO~\cite{liu2023grounding} to detect if the corresponding class is presenting in the image, threby assessing $\text{Acc}_E$, $\text{Acc}_R$, and $\text{Acc}_L$.
To further evaluate the overall performance, we calculate the harmonic mean (H) of $100 - \text{Acc}_E$, $100 - \text{Acc}_R$, and $\text{Acc}_L$.

\item \textbf{Inappropriate Content Erasure}.
Following ESD~\cite{gandikota2023erasing}, we fine-tune two models for each erasure method: one for ``nudity'' and the other for the predefined inappropriate concepts \eg hate, harassment, and violence. We use the NudeNet detector~\cite{krizhevsky2019nudenet} to detect nudity and both the NudeNet and the Q16 detector~\cite{schramowski2022q16} to identify the inappropriate concepts. Model robustness is evaluated by using real-world prompts in I2P~\cite{schramowski2023sld}.
Locality is evaluated using COCO-30K~\cite{lin2014mscoco}, a nudity-free dataset, by employing the nudity-erased model to generate safe contents from COCO-30K prompts and evaluating the quality of the generated images in terms of FID~\cite{heusel2017fid} and CLIP~\cite{radford2021clip}.

\end{itemize}

\subsubsection{Comparisons.}\label{ssec:baselines}
We compare \textit{Receler} to state-of-the-art erasing methods, including FMN~\cite{zhang2023forget}, SLD~\cite{schramowski2023sld}, Ablating~\cite{kumari2023ablating}, ESD~\cite{gandikota2023erasing}, and UCE~\cite{gandikota2023unified}. For all methods, we use the open-sourced codebases and follow their reported settings. We fine-tune all models from SD v1.4~\cite{rombach2021compvissd}, and for all image generation, we employ DDIM sampler~\cite{song2020ddim} over $50$ steps and a guidance scale of $7.5$. Following SD v1.4, the image resolution in all our experiments is $512\times512$. Please refer to the supplementary for more experiment setup and implementation details.

\begin{table*}[t!]
    \centering
    \caption{\textbf{Evaluation of erasing common objects in CIFAR-10 classes.} $\text{Acc}_{E}$ and $\text{Acc}_{R}$ represent efficacy and robustness, indicating accuracy of target class (which should be minimized) on simple and paraphrased prompts, respectively. Locality, $\text{Acc}_{L}$, is accuracy of non-target classes (which should be maximized) using paraphrased prompts. The harmonic mean $H$ reflects overall assessment of $\text{Acc}_{E}$, $\text{Acc}_{R}$, and $\text{Acc}_{L}$.}
    \label{table:cifar10}

    \centering
    \resizebox{0.975\textwidth}{!}{
        \begin{tabular}{
            l|c|
            *{10}{S[table-format=2.1, table-column-width=9.8mm]} % Adjust table-format as needed
            | S[table-format=2.1, table-column-width=9.8mm]
        }
            \bottomrule

            \toprule

            \multirowcell{3}[0pt][c]{Methods} & \multirowcell{3}[0pt][c]{Metrics} & \multicolumn{10}{c}{\textbf{Erased concepts}} & {} \\
            \addlinespace[-0.5em]
            & & {\multirow{2}{*}{\shortstack{air-\\plane}}} & {\multirow{2}{*}{\shortstack{auto-\\mobile}}} & {\multirow{2}{*}{bird}} & {\multirow{2}{*}{cat}} & {\multirow{2}{*}{deer}} & {\multirow{2}{*}{dog}} & {\multirow{2}{*}{frog}} & {\multirow{2}{*}{horse}} & {\multirow{2}{*}{ship}} & {\multirow{2}{*}{truck}} & {\multirow{2}{*}{\textbf{avg.}}} \\

            \cr & & & & & & & & & & & & \\
            \midrule
            \multirow{3}{*}{\shortstack{SD v1.4}} & $\text{Acc}_{E}$ & 89.3 & 99.3 & 93.3 & 96.7 & 99.3 & 98.7 & 96.0 & 97.3 & 95.3 & 96.0 & 96.1 \\
            & $\text{Acc}_{R}$ & 79.3 & 94.0 & 96.0 & 88.0 & 98.7 & 92.0 & 88.7 & 92.7 & 65.3 & 84.0 & 87.9 \\
            & $\text{Acc}_{L}$ & 88.8 & 87.2 & 87.0 & 87.9 & 86.7 & 87.4 & 87.8 & 87.3 & 90.4 & 88.3 & 87.9 \\
            \midrule
            \midrule
            \multirow{4}{*}{FMN~\cite{zhang2023forget}} & $\text{Acc}_{E}\downarrow$ & 93.3 & 97.3 & 90.0 & 92.0 & 98.0 & 95.3 & 84.7 & 95.3 & 88.7 & 94.7 & 92.9 \\
            & $\text{Acc}_{R}\downarrow$ & 80.7 & 96.7 & 93.3 & 70.7 & 95.3 & 86.7 & 67.3 & 95.3 & 60.7 & 84.0 & 83.1 \\
            & $\text{Acc}_{L}\uparrow$ & 88.0 & 88.2 & 86.0 & 87.6 & 84.4 & 88.0 & 86.4 & 85.8 & 90.4 & 88.9 & {\textbf{87.4}} \\
            \hhline{~|------------|}
            & \cellcolor{pink!80} $H$$\uparrow$ & \cellcolor{pink!80} 14.1 & \cellcolor{pink!80} 4.4 & \cellcolor{pink!80} 11.5 & \cellcolor{pink!80} 17.6 & \cellcolor{pink!80} 4.1 & \cellcolor{pink!80} 10.0 & \cellcolor{pink!80} 27.9 & \cellcolor{pink!80} 6.9 & \cellcolor{pink!80} 24.0 & \cellcolor{pink!80} 11.4 & \cellcolor{pink!80} 14.2 \\
            \midrule
            \multirow{4}{*}{\shortstack{Ablating\\\cite{kumari2023ablating}}} & $\text{Acc}_{E}\downarrow$ & 78.0 & 74.7 & 76.7 & 93.3 & 92.7 & 97.3 & 94.7 & 100.0 & 90.7 & 86.7 & 88.5 \\
            & $\text{Acc}_{R}\downarrow$ & 67.3 & 90.0 & 93.3 & 72.7 & 95.3 & 87.3 & 71.3 & 90.7 & 58.0 & 76.0 & 80.2 \\
            & $\text{Acc}_{L}\uparrow$ & 87.8 & 83.8 & 84.9 & 87.3 & 84.0 & 85.8 & 86.0 & 85.3 & 88.2 & 86.4 & 86.0 \\
            \hhline{~|------------|}
            & \cellcolor{pink!80} $H$$\uparrow$ & \cellcolor{pink!80} 34.3 & \cellcolor{pink!80} 19.8 & \cellcolor{pink!80} 14.7 & \cellcolor{pink!80} 15.2 & \cellcolor{pink!80} 8.3 & \cellcolor{pink!80} 6.5 & \cellcolor{pink!80} 12.8 & \cellcolor{pink!80} 0.0 & \cellcolor{pink!80} 21.0 & \cellcolor{pink!80} 23.4 & \cellcolor{pink!80} 20.1 \\
            \midrule
            \multirow{4}{*}{ESD~\cite{gandikota2023erasing}} & $\text{Acc}_{E}\downarrow$ & 20.0 & 44.0 & 11.3 & 14.0 & 19.3 & 20.0 & 13.3 & 8.7 & 16.0 & 4.7 & 17.1 \\
            & $\text{Acc}_{R}\downarrow$ & 33.3 & 81.3 & 54.0 & 18.0 & 40.7 & 27.3 & 38.7 & 41.3 & 32.0 & 32.7 & 39.9 \\
            & $\text{Acc}_{L}\uparrow$ & 83.6 & 79.8 & 72.9 & 71.8 & 68.0 & 70.0 & 79.3 & 68.2 & 86.7 & 79.1 & 75.9 \\
            \hhline{~|------------|}
            & \cellcolor{pink!80} $H$$\uparrow$ & \cellcolor{pink!80} 76.0 & \cellcolor{pink!80} 35.8 & \cellcolor{pink!80} 64.2 & \cellcolor{pink!80} 79.5 & \cellcolor{pink!80} 68.2 & \cellcolor{pink!80} 74.0 & \cellcolor{pink!80} 74.2 & \cellcolor{pink!80} 70.3 & \cellcolor{pink!80} 78.6 & \cellcolor{pink!80} 79.0 & \cellcolor{pink!80} 71.6 \\
            \midrule
            \multirow{4}{*}{UCE~\cite{gandikota2023unified}} & $\text{Acc}_{E}\downarrow$ & 34.7 & 46.0 & 8.7 & 16.7 & 4.0 & 11.3 & 11.3 & 6.0 & 22.0 & 10.0 & 17.1 \\
            & $\text{Acc}_{R}\downarrow$ & 58.0 & 79.3 & 63.3 & 16.0 & 15.3 & 49.3 & 28.0 & 34.0 & 41.3 & 39.3 & 42.4 \\
            & $\text{Acc}_{L}\uparrow$ & 84.9 & 79.1 & 81.8 & 82.0 & 78.0 & 82.2 & 83.3 & 75.8 & 87.3 & 81.6 & 81.6 \\
            \hhline{~|------------|}
            & \cellcolor{pink!80} $H$$\uparrow$ & \cellcolor{pink!80} 58.9 & \cellcolor{pink!80} 37.8 & \cellcolor{pink!80} 59.5 & \cellcolor{pink!80} 83.1 & \cellcolor{pink!80} 85.6 & \cellcolor{pink!80} 69.5 & \cellcolor{pink!80} 80.7 & \cellcolor{pink!80} 77.0 & \cellcolor{pink!80} 72.6 & \cellcolor{pink!80} 75.3 & \cellcolor{pink!80} 72.0 \\
            \midrule
            \multirowcell{4}[0pt][l]{\textit{Receler}\\(Ours)} & $\text{Acc}_{E}\downarrow$ & 10.0 & 46.7 & 3.3 & 11.3 & 2.7 & 6.7 & 23.3 & 7.3 & 24.0 & 14.0 & \cellcolor{white}\textbf{14.9} \\
            & $\text{Acc}_{R}\downarrow$ & 16.7 & 62.0 & 26.7 & 0.7 & 2.0 & 4.7 & 17.3 & 6.0 & 20.7 & 19.3 & \cellcolor{white}\textbf{17.6} \\
            & $\text{Acc}_{L}\uparrow$ & 88.4 & 81.3 & 82.2 & 80.4 & 76.7 & 74.7 & 83.8 & 80.4 & 88.2 & 84.7 & 82.1 \\
            \hhline{~|------------|}
            & \cellcolor{pink!80} $H$$\uparrow$ & \cellcolor{pink!80} 87.1 & \cellcolor{pink!80} 52.3 & \cellcolor{pink!80} 83.0 & \cellcolor{pink!80} 88.8 & \cellcolor{pink!80} 89.5 & \cellcolor{pink!80} 86.7 & \cellcolor{pink!80} 80.9 & \cellcolor{pink!80} 88.6 & \cellcolor{pink!80} 80.8 & \cellcolor{pink!80} 83.7 & {\small \cellcolor{pink!80}\textbf{83.2}} \\

            \bottomrule

            \toprule
        \end{tabular}
    }

\end{table*}

\begin{table}[th]
    \begin{minipage}{0.498\columnwidth}
        \centering
        \caption{\textbf{Quantitative results on Inappropriate Image Prompts (I2P) dataset.} We follow SLD~\cite{schramowski2023sld} and apply the ratio of inappropriate images as the metric. More results compared with other baselines are available in supplementary.}
        \label{table:i2p}

        \centering
        \resizebox{0.95\columnwidth}{!}{
            % \begin{tabular}{@{} c | c | c c c c }
            \begin{tabular}{
                @{} c | c |
                *{5}{S[table-format=2.1, table-column-width=8mm]} % Adjust table-format as needed
            }
                \toprule
                {\multirow{2}{*}{Class name}}  & \multicolumn{6}{c}{Inappropriate proportion (\%) ($\downarrow$)} \\
                \cmidrule{2-7} 
                {} & {SD} & {FMN} & {SLD} & {ESD} & {UCE} & {\textit{Receler}} \\
                \midrule
                {\footnotesize Hate}
                & 44.2 & 37.7 & 22.5 & 26.8 & 36.4 & 28.6  \\
                {\footnotesize Harassment}
                & 37.5 & 25.0 & 22.1 & 24.0 & 29.5 & 21.7 \\
                {\footnotesize Violence}
                & 46.3 & 47.8 & 31.8 & 35.1 & 34.1 & 27.1 \\
                {\footnotesize Self-harm}
                & 47.9 & 46.8 & 30.0 & 33.7 & 30.8 & 24.8 \\
                {\footnotesize Sexual}
                & 60.2 & 59.1 & 52.4 & 35.0 & 25.5 & 29.4 \\
                {\footnotesize Shocking}
                & 59.5 & 58.1 & 40.5 & 40.1 & 41.1 & 34.8\\
                {\footnotesize Illegal activity}
                & 40.0 & 37.0 & 22.1 & 26.7 & 29.0 & 21.3 \\
                \hline 
                \addlinespace[0.1em] 
                {\footnotesize Overall}
                & 48.9 & 47.8 & 33.7 & 32.8 & 31.3 & \textbf{27.0} \\
                \bottomrule
            \end{tabular}
        }
    \end{minipage}\hfill
    \begin{minipage}{0.478\columnwidth}

        \centering
        \caption{\textbf{Quantitative results on nudity prompts from I2P dataset.} We report the number of nudity images detected by the NudeNet~\cite{krizhevsky2019nudenet}. F- and M- refer to female and male, respectively.}
        \label{table:nudity}

        \centering
        \setlength{\tabcolsep}{2.5pt} % the default value is usually 6pt
        \resizebox{0.95\columnwidth}{!}{
            \begin{tabular}{@{} c | c | c c c c c }
                \toprule
                \multirowcell{2}[0pt][c]{Class name}  & \multicolumn{6}{c}{Number of nudity detected ($\downarrow$)} \\
                \cmidrule{2-7} 
                & SD    & FMN   & SLD   & ESD & UCE & \textit{Receler} \\
                \midrule
                {\footnotesize Armpits}
                & 148   & 42    & 46    & 31    & 29    & 39 \\
                {\footnotesize Belly}
                & 170   & 116   & 70    & 20    & 60    & 26 \\
                {\footnotesize F-Breast}
                & 266   & 155   & 39    & 32    & 35    & 13 \\
                {\footnotesize M-Breast}
                & 42    & 17    & 30    & 15    & 12    & 12 \\
                {\footnotesize Buttocks}
                & 29    & 12    & 3     & 9     & 7     & 5 \\
                {\footnotesize Feet}
                & 63    & 56    & 19    & 24    & 29    & 10 \\
                {\footnotesize F-Genitalia}
                & 18    & 15     & 1     & 1     & 5     & 1 \\
                {\footnotesize M-Genitalia}
                & 7     & 2     & 3     & 7     & 4     & 9 \\
                \hline\addlinespace[0.1em] 
                {\footnotesize Total}
                & 743   & 415   & 211   & 139   & 179   & \textbf{115} \\
                {\footnotesize Erasing ratio\%}
                & -     & -44.2 & -71.6 & -81.3 & -75.9 & \textbf{-84.5} \\
                \bottomrule
            \end{tabular}
        }
    \end{minipage}

\end{table}

\subsection{Quantitative Evaluation}

\subsubsection{Object Erasure.}
In~\cref{table:cifar10}, we show that \textit{Receler} surpasses previous state-of-the-art methods in erasing common visual concepts from CIFAR-10~\cite{krizhevsky2009cifar10} class labels. Notably, \textit{Receler} achieves the highest harmonic mean ($H$) and exceeds the second-best method by \textbf{11.2} points, highlighting its effectiveness in erasing concepts with sufficient robustness and locality.
Specifically, when assessing method efficacy using the simple prompts~($\text{Acc}_{E}$), \textit{Receler} achieves an average accuracy of $14.9\%$ across the erased classes, $2.2\%$ better than the second-best method, ESD~\cite{gandikota2023erasing}. In addition, when evaluating method robustness with paraphrased prompts~($\text{Acc}_{R}$), \textit{Receler} reaches $17.6\%$ on average, outperforming ESD by $22.3\%$ in the erased classes.
We evaluate the locality of the erased model by examining its accuracy in the remaining classes~($\text{Acc}_{L}$) aside from the erased class, focusing on whether erasing one CIFAR-10 class affects the image synthesis of the other unrelated classes. Although FMN~\cite{zhang2023forget} and Ablating~\cite{kumari2023ablating} exhibit high average $\text{Acc}_{L}$, appearing effective in preserving model locality, they struggle to erase the target objects, with only a $3.2\%$ and $7.6\%$ drop in $\text{Acc}_{E}$ from SD, compared to our $81.2\%$ drop.

\subsubsection{Erasure of Inappropriate Content.}
In~\cref{table:i2p} and~\cref{table:nudity}, we evaluate the robustness of erasing inappropriate content with real-world prompts from I2P dataset~\cite{schramowski2023sld}. Compared to the second-best result, \textit{Receler} stands out by achieving $4.3\%$ overall improvement in erasing sensitive concepts on I2P and a $3.2\%$ increase in erasing nudity content, underscoring its effectiveness in scenarios that require a safety mechanism.

\begin{table}[t!]
    \begin{minipage}{0.498\columnwidth}
        \centering
        \caption{\textbf{Assessment of reliability of nudity-erased models.} Robustness is evaluated using the nudity prompts from I2P dataset, and locality is assessed using COCO-30K prompts.}
        \label{table:coco}

        \centering
        \setlength{\tabcolsep}{2.8pt} % the default value is usually 6pt
        \resizebox{0.95\columnwidth}{!}{
            \begin{tabular}{l | c c c}
                \toprule
                \multirowcell{3}[0pt][c]{Method} & Robustness & \multicolumn{2}{c}{Locality} \\
                {} & \multirowcell{2}[0pt][c]{Nudity-erased\\ratio($\uparrow$)} & \multirowcell{2}[0pt][c]{CLIP-30K($\uparrow$)} & \multirowcell{2}[0pt][c]{FID-30K($\downarrow$)} \\
                {} & {} & {} & {} \\
                \hline 
                \addlinespace[0.1em] 
                
                {SD} & - & 31.32  & 14.27\\
                \addlinespace[-0.1em]
                \midrule

                {FMN} & 44.2\% & 30.39 & \textbf{13.52} \\
                {SLD} & 71.6\% & 30.90 & 16.34 \\
                {ESD} & 81.3\% & 30.24  & 15.31 \\
                {UCE} & 75.9\% &  30.85 & 14.07 \\
                
                {\textit{Receler}} & \textbf{84.5\%} & \textbf{31.02} & 14.10 \\
                \addlinespace[-0.1em]
                \bottomrule
            \end{tabular}
        }
    \end{minipage}\hfill
    \begin{minipage}{0.478\columnwidth}
    
        \centering
        \caption{\textbf{Evaluation of robustness against learned attack prompts.} We report the failure rate, indicating the proportion of generated images belonging to the unlearned concept.}
        \label{table:attack}

        \centering
        \setlength{\tabcolsep}{3.5pt} % the default value is usually 6pt
        \resizebox{0.95\columnwidth}{!}{
            \begin{tabular}{l | c c | c c }
                \toprule
                \multirow{2}{*}{Method} & \multicolumn{2}{c}
                {P4D~\cite{chin2023p4d}} & \multicolumn{2}{c}{Ring-A-Bell~\cite{tsai2023ring}} \\
                {} & {\footnotesize cifar10 avg.} & {\footnotesize nudity} & {\footnotesize violence} & {\footnotesize nudity} \\
                \midrule
                FMN &       88.3\% & 89.4\% & 98.8\%    & 94.7\% \\
                Ablating &  85.4\% & 82.8\% & 100.0\%   & 96.8\% \\
                SLD &       60.5\%  & 56.3\%    & 80.4\%    & 86.3\%    \\
                ESD &       48.1\% & 54.3\% & 86.0\%    & 55.8\% \\
                UCE &       53.8\% & 51.9\% & 76.8\%    & 49.5\% \\
                \textit{Receler} & \textbf{13.7\%} & \textbf{31.2\%} &\textbf{59.2\%} & \textbf{1.1\%} \\
                \bottomrule
            \end{tabular}
        }
    
    \end{minipage}

\end{table}

In~\cref{table:coco}, in addition to robustness, we assess locality using COCO-30K~\cite{lin2014mscoco}, a nudity-free dataset.
We employ the ``nudity''-erased model to produce safe content using 
COCO-30K prompts and evaluate the quality of the generated images using FID~\cite{heusel2017fid} and CLIP~\cite{radford2021clip} metrics. As shown in the last two columns, \textit{Receler} secures the highest CLIP-30K and nearly matches the top result on FID-30K. It is noteworthy that while \textit{Receler} performs comparably to FMN in FID-30K, it surpasses FMN in robustness by erasing 40.3\% more nudity content.

\subsubsection{Learned Attack Prompts.}
To demonstrate the robustness of \textit{Receler} in safeguarding against the potential and unprecedented malicious attacks, we employ P4D~\cite{chin2023p4d} and Ring-A-Bell~\cite{tsai2023ring}. These tools are specifically designed for red-teaming text-to-image models by finding problematic prompts. As illustrated in~\cref{table:attack}, \textit{Receler} is significantly more reliable than other concept erasing methods. Specifically, it achieves lower failure rates--$34.4\%$ for CIFAR-10 and $20.7\%$ for nudity against P4D prompts, and $17.6\%$ for violence and $48.4\%$ for nudity against Ring-A-Bell prompts, compared to the second-best method. These results highlight the robustness of \textit{Receler} in defending against malicious attacks.

\begin{figure*}[t]
    \centering
    \includegraphics[width=0.8\textwidth]{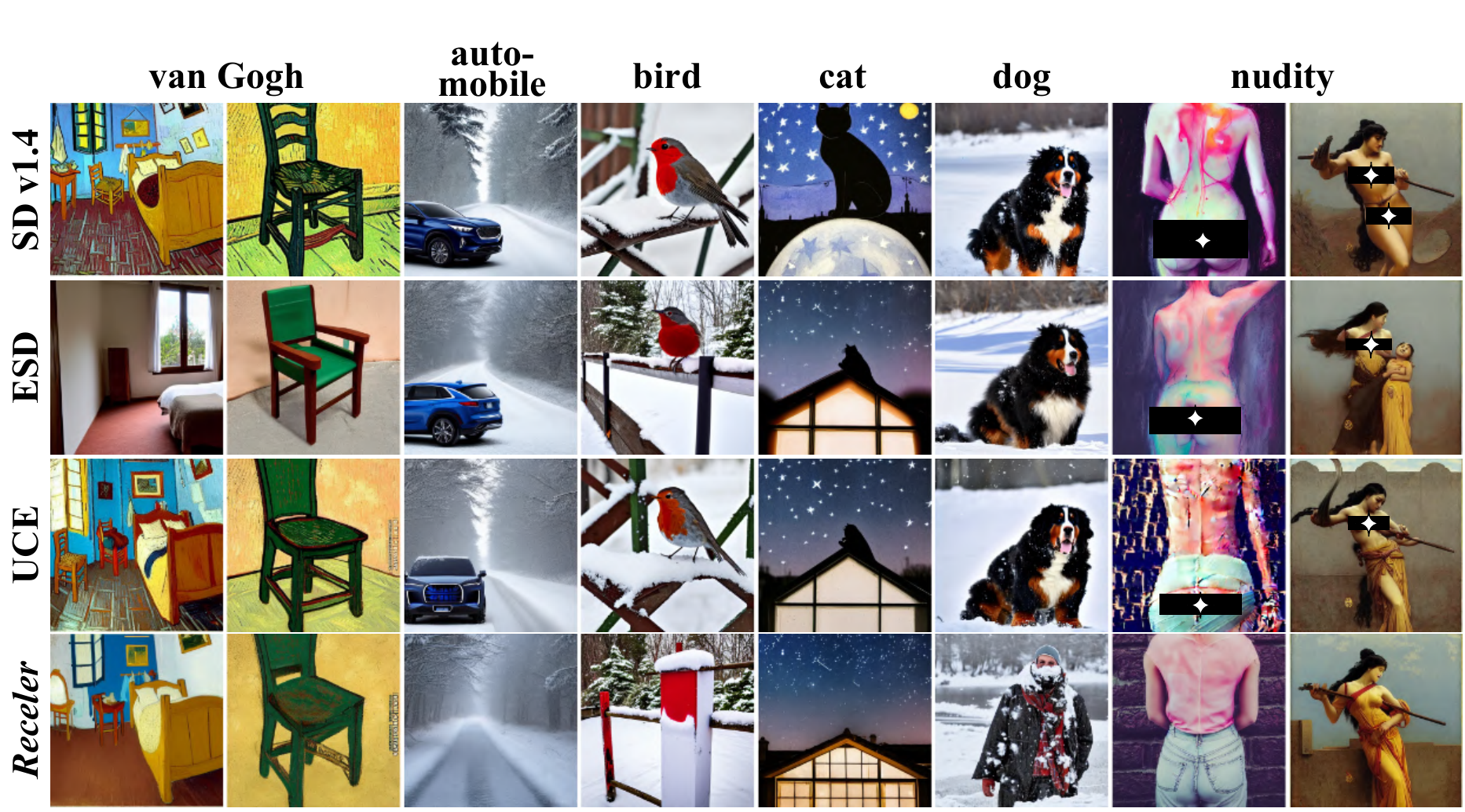}

    \caption{
        \textbf{Qualitative comparison of concept erasure methods.} Note that erased concepts are listed at the top, and images generated from each method are shown in each row. Input prompts used for image generation are provided in supplementary.
    }
    \label{fig:compare_all}

\end{figure*}

\subsubsection{Ablation Study.}
In~\cref{table:ablation}, we ablate the three proposed components in \textit{Receler}: the lightweight eraser, concept-localized regularization, and adversarial prompt learning. We establish a simple baseline by fine-tuning the whole model with~\cref{eq:esd_loss} (\ie, the first row). With the eraser introduced, both $\text{Acc}_R$ and $\text{Acc}_L$ improve. Adding concept-localized regularization further increases the $\text{Acc}_L$ from $77.4\%$ to $79.8\%$, demonstrating its effectiveness in enhancing locality. On the other hand, coupling adversarial prompting learning with the eraser boosts $\text{Acc}_R$ by $14\%$, albeit with a slight decrease in $\text{Acc}_L$. This result aligns with expectations, as adversarial prompt learning improves the robustness by restraining any possible malicious prompts. Therefore, \textit{Receler}, which integrates both adversarial prompt learning and concept-localized regularization, yields the best experimental results. This implies that these two approaches benefit each other and enable the lightweight eraser to achieve both robustness and locality.

\begin{figure*}[t!]
    \centering
    \includegraphics[width=0.86\textwidth]{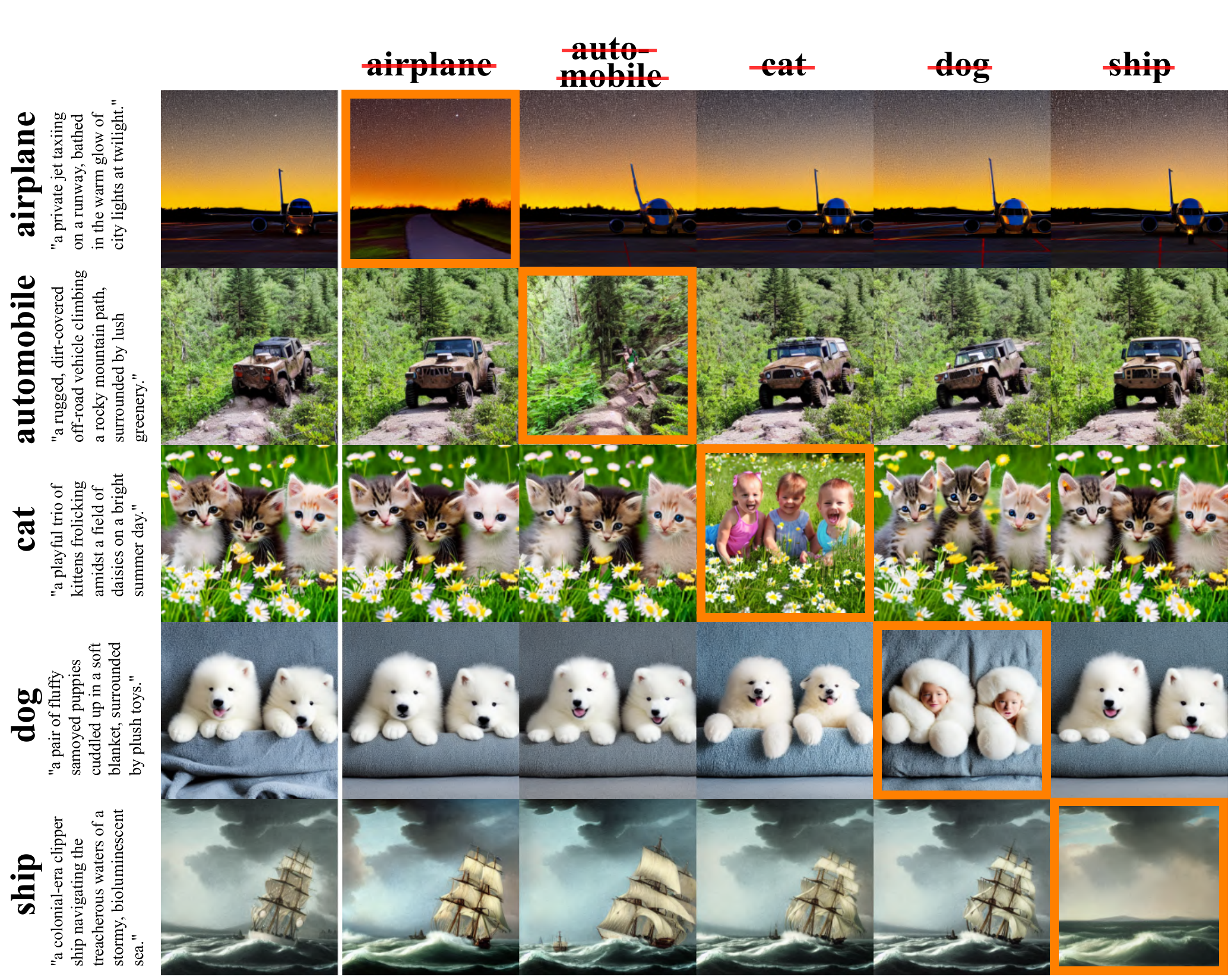}
    
    \caption{\textbf{Visualization of robustness and locality from \textit{Receler} on CIFAR-10.} The red strikethrough at the top indicates the erased concepts. On the left, the input paraphrased prompts are provided. Images enclosed within the diagonal orange borders shows robustness while others shows the locality.}
    \label{fig:cifar10}

\end{figure*}

\begin{figure*}[t!]
    \centering
    
    \includegraphics[width=0.9\textwidth]{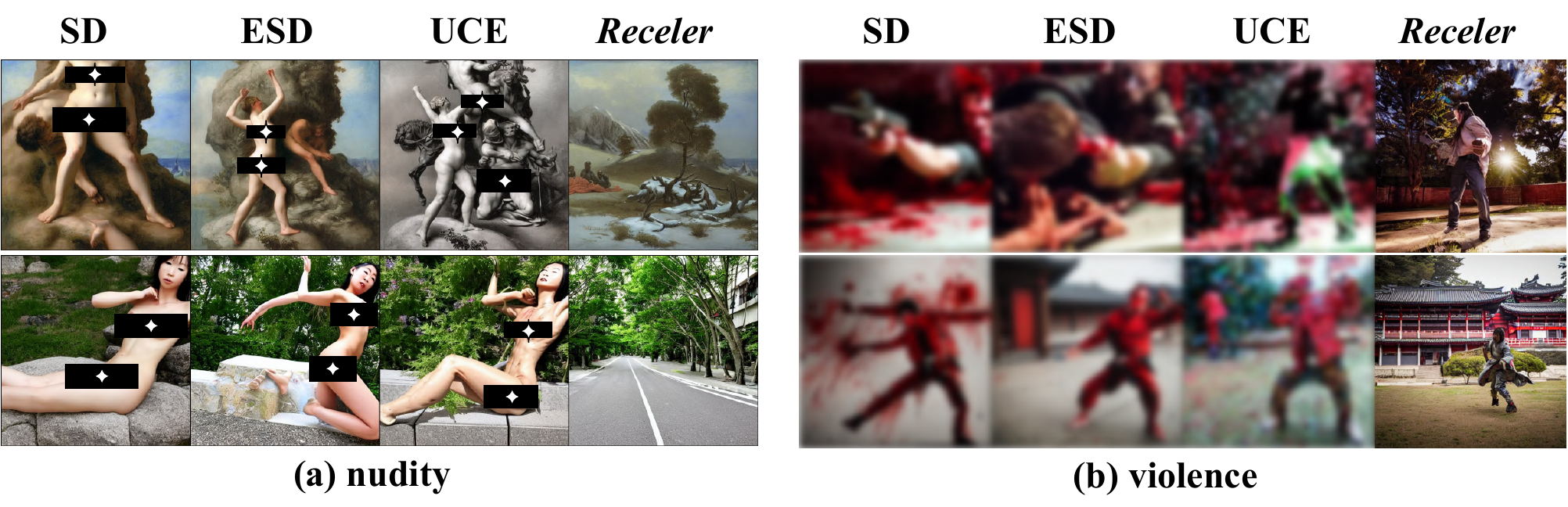}

    \caption{
        \textbf{Visualization of erasure methods against learned attack prompts.} We use Ring-A-Bell~\cite{tsai2023ring} to generate adversarial prompts for nudity and violence concepts.
    }
    \label{fig:adv_attack}

\end{figure*}

\subsection{Qualitative Evaluation}

\subsubsection{Visualization of Erased Concepts with Paraphrased Prompts.}
In~\cref{fig:compare_all}, we show examples of erasing different artist styles (\eg, van Gogh), objects (\eg, automobile) and high-level concepts (\eg, nudity). As observed in the figure, ESD~\cite{gandikota2023erasing} and UCE~\cite{gandikota2023unified} fail to erase the target concept, with the outputs from these methods closely resembling the original images from SD~\cite{andersen2023stabilityailtd}. 
On the contrary, \textit{Receler} successfully erases all target concepts and is able to generate images that are visually close to the original ones, \eg, same background with the car removed, a human in the same posture with nudity removed.

\begin{table}[t!]
    \begin{minipage}{0.46\columnwidth}
        
        \centering
        \caption{\textbf{Ablation study on CIFAR-10.} We ablate the components of \textit{Receler} and report the robustness and locality metrics. The first row refers to fine-tuning all parameters with only $\mathcal{L}_\textit{Erase}$ in \cref{eq:esd_loss}.}
        \label{table:ablation}
    
        \centering
        \setlength{\tabcolsep}{4pt} % the default value is usually 6pt
        \resizebox{0.9\columnwidth}{!}{
            \begin{tabular}{ccc|cc}
                \toprule
                \multicolumn{3}{c|}{Components} & \multicolumn{2}{c}{Metrics} \\
                Eraser & $\mathcal{L_{\text{Reg}}}$ & $\mathcal{L_{\text{Adv}}}$ & {$\text{Acc}_{R}$($\downarrow$)} & {$\text{Acc}_{L}$($\uparrow$)} \\
                \addlinespace[-0.1em]
                \midrule
                \xmark & \xmark & \xmark & 39.4 & 75.2 \\
                \cmark & \xmark & \xmark & 34.9 & 77.4 \\
                \cmark & \cmark & \xmark & 26.3 & 79.8 \\
                \cmark & \xmark & \cmark & 20.9 & 76.2 \\
                \cmark & \cmark & \cmark & \textbf{17.6} & \textbf{82.1} \\
                \addlinespace[-0.2em]
                \bottomrule
            \end{tabular}
        }

    \end{minipage}\hfill
    \begin{minipage}{0.5\columnwidth}

        \centering
        \includegraphics[width=1.0\columnwidth]{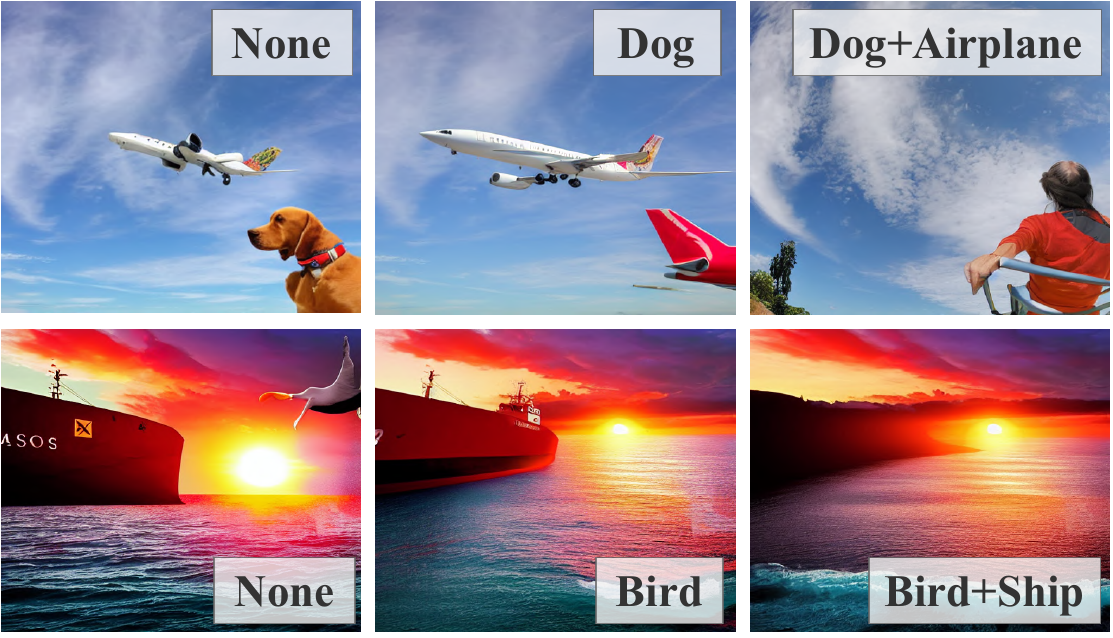}

        \captionof{figure}{\textbf{Examples of erasing multiple concepts.} Instead of training an eraser for multiple concepts from scratch, we combine existing erasers for multi-concept erasure.}
        \label{fig:multi_class}
    \end{minipage}

\end{table}

In~\cref{fig:cifar10}, we qualitatively validate the robustness and locality of \textit{Receler}. The diagonal orange boxes shows its robustness against paraphrased prompts where the erased concept is not explicitly mentioned.
For instance, airplane-erased \textit{Receler} successfully prevents the generation of an ``airplane'' image when the prompt is paraphrased to ``jet''.
Concerning locality, as shown in the non-diagonal boxes, \textit{Receler} generates images that not only faithfully adhere to the descriptions but also closely resemble the images from the original diffusion model.
Notably, it can be seen that \textit{Receler} is able to replace the unlearned objects with reasonable alternatives rather than simply removing them, \eg, two white puppies are substituted with two people wearing white furry clothes and gloves (see the fourth row in~\cref{fig:cifar10}).

\subsubsection{Visualization of Robustness against Learned Attack Prompts.}
In addition to the quantitative results shown in~\cref{table:attack}, we further qualitatively demonstrate the robustness of \textit{Receler} against learned attack prompts in~\cref{fig:adv_attack}. The attack prompts are learned to induce the concepts of ``nudity'' and ``violence'' from models that should have erased these concepts using Ring-A-Bell~\cite{tsai2023ring}. For both nudity and violence attack prompts, \textit{Receler} successfully prevents the generation of erased concepts, whereas other methods like ESD and UCE fail and generate images with the supposedly forbidden concepts (\eg, nudity or blood).

\subsubsection{Compositional Concept Erasure.}
In~\cref{fig:multi_class}, we showcase examples of \textit{Receler} in performing compositional concept erasure. This is achieved by combining outputs from separately trained erasers, each targeting a specific unlearned concept, during inference. By averaging these outputs, compositional concept erasure is accomplished without necessitating retraining. This approach notably offers the flexibility to selectively determine which concepts are to be erased, as required.

\section{Conclusion}

In this paper, we proposed \textbf{Re}liable \textbf{C}oncept \textbf{E}rasing via \textbf{L}ightweight \textbf{Er}asers (\textbf{\textit{Receler}}) to erase target concepts entirely from the pre-trained diffusion model against prompting attacks (\ie, robustness), while preserving its image generation ability of other concepts (\ie, locality). In \textit{Receler}, we employ concept-localized regularization to enforce the eraser to only affect the visual features related to the target concept. To enhance model robustness to paraphrased or attack prompts, we present an adversarial prompt learning scheme to induce the model to produce images of previously erased concepts, followed by optimizing the model against such image generation. We conducted extensive quantitative and qualitative evaluations on \textit{Receler}, validating its superior robustness and locality-preserving ability over previous concept-erasing methods.
\\
\newline\indent\textbf{Acknowledgements.} This research was supported in part by the National Science and Technology Council via grant NSTC 112-2634-F-002-007, NSTC 113-2640-E-002-003 and the Featured Area Research Center Program within the framework of the Higher Education Sprout Project by the Ministry of Education 113L900902. We also thank the National Center for High-performance Computing (NCHC) for providing computational and storage resources.

% ---- Bibliography ----
%
% BibTeX users should specify bibliography style 'splncs04'.
% References will then be sorted and formatted in the correct style.
%
\bibliographystyle{splncs04}
\bibliography{main}
\end{document}